\documentclass[conference]{IEEEtran}
\renewcommand\IEEEkeywordsname{Keywords}
\usepackage{cite}
\usepackage{amsmath,amssymb,amsfonts}
\usepackage{algorithmic}
\usepackage{graphicx}
\usepackage{textcomp}
\usepackage{xcolor}
\usepackage{float}
\usepackage{authblk}
\usepackage[hidelinks]{hyperref}
\UseRawInputEncoding
\def\BibTeX{{\rm B\kern-.05em{\sc i\kern-.025em b}\kern-.08em
    T\kern-.1667em\lower.7ex\hbox{E}\kern-.125emX}}

\usepackage{amsmath}
\usepackage{graphicx}

\makeatletter

\def\ps@IEEEtitlepagestyle{%
  \def\@oddfoot{\mycopyrightnotice}%
  \def\@evenfoot{}%
}
\def\mycopyrightnotice{%
  {\footnotesize 978-1-6654-6316-4/22/\$31.00~\copyright~2022 IEEE\hfill}
  \gdef\mycopyrightnotice{}
}

\@ifundefined{showcaptionsetup}{}{
 \PassOptionsToPackage{caption=false}{subfig}}
\usepackage{subfig}
\makeatother

\usepackage{eso-pic}

\title{Gastrointestinal Disorder Detection with a Transformer Based Approach\\
}

\begin{document}



\author[ 1]{\rm A.K.M. Salman Hosain}
\author[ 1]{\rm Mynul islam}
\author[1 ]{\rm Md Humaion Kabir Mehedi}

\author[ 2]{\rm Irteza Enan Kabir}
\author[ 3]{\rm \\Zarin Tasnim Khan }

\affil[1 ]{Department of Computer Science and Engineering, Brac University, Dhaka, Bangladesh }

\affil[ 2]{University of Rochester, NY, United States }
\affil[3 ]{Shaheed Tajuddin Ahmad Medical College, Dhaka, Bangladesh}

\affil[1 ]{\textit { \{akm.salman.hosain, mynul.islam, humaion.kabir.mehedi \}@g.bracu.ac.bd}}
\affil[2,3 ]{\textit {\{irtezaenan, zarinkhan27\}@gmail.com}}

\maketitle

\begin{abstract}
Accurate disease categorization using endoscopic images is a significant problem in Gastroenterology. This paper describes a technique for assisting medical diagnosis procedures and identifying gastrointestinal tract disorders based on the categorization of characteristics taken from endoscopic pictures using a vision transformer and transfer learning model. Vision transformer has shown very promising results on difficult image classification tasks. In this paper, we have suggested a vision transformer based approach to detect gastrointestianl diseases from wireless capsule endoscopy (WCE) curated images of colon with an accuracy of 95.63\%. We have compared this transformer based approach with pretrained convolutional neural network (CNN) model DenseNet201 and demonstrated that vision transformer surpassed DenseNet201 in various quantitative performance evaluation metrics.

\end{abstract}


\section{Introduction}
The gastrointestinal (GI) tract, also known as digestive tract is prone to several diseases such as polyps, ulcer, colorectal cancer, etc ~\cite{Haile_Salau_Enyew_Belay_2022}. Common symptoms include pain or discomfort in the abdomen, loss of appetite, nausea and vomiting, abdominal discomfort and fatigue. Some of the GI diseases often lead to GI cancer, which is considered the second most common cancer worldwide ~\cite{Sitarz_Skierucha_Mielko_Offerhaus_Maciejewski_Polkowski_2018}. One of the common diseases of the gastro-intenstine is the muco-submucosal polyps, which are the results of chronic prolapse of the mucosa in intestine. ~\cite{Tan_2013}. Polyps often don't show a lot of symptoms in the early stages, but as it enlarges, it can block the opening to the small intestine. The symptoms for polyps might include blood in stool thus anemia, tenderness when the stomach is touched and nausea. These appear as polypoid mass in endoscopic imaging, and has an increased risk of cancer. Esophagitis is another common GI condition which is caused from the inflammation of the tube connecting the throat to the stomach. Esophagitis mainly causes difficulties in swallowing, chest pain, heart burn, swallowed food being stuck in esophagus ~\cite{Moayyedi_Santana_Khan_Preston_Donnellan_2007}. Endoscopy usually shows rings of abnormal tissue. Ulcerative colitis, an inflammatory bowel disease, is also a frequently occurring condition, which causes inflammation in the GI tract along with abdominal pain, diarrhoea, fatigue and bloody stool. 

These GI diseases often have overlapping symptoms, thus difficult to identify. Initial diagnosis of these diseases may lead to cure or prevention from developing fatal cancer. Although visual assessment of endoscopy images give an initial diagnosis, this is often time consuming and highly subjective ~\cite{JHA2021102007}. Moreover, there might be radiologist deficiencies and other human factors which often lead to false positive or even false negative diagnosis, which can be detrimental for the patient ~\cite{Ozturk_Ozkaya_2021}. Thus, a computer aided diagnosis would be valuable for high accuracy detection at the early stages. 

In this paper, we classify endoscopic images for subjects with gastrointestinal diseases. For the classification task, we undertook two different approaches. We used vision transformer and transfer learning method with pretrained CNN architecture for the classification, and compared the results between these the two classification models. The gastrointestinal diseases for our data set consists of four classes:

\begin{itemize}
    \item Healthy control, or normal class
    \item Ulcerative colitis
    \item Polyps and
     \item Esophagitis
\end{itemize}

Our contributions in this work are - 

\begin{itemize}
    \item We have utilized vision transformer based model (ViT) and pretrained CNN model DenseNet201 to detect three gastrointestinal diseases along with healthy colon from wireless capsule endoscopy images
    (WCE) curated images of colon
    \item We have conducted comparative analysis between the two models on various quantitative performance evaluation metrics and demonstrated the superior classifier
\end{itemize}

\section{Related Works}

Machine learning techniques have been previously used in the area of medicine for diagnosis purposes, such as using neural networks for classification of stomach cancer ~\cite{Majid_Khan_Yasmin_Rehman_Yousafzai_Tariq_2020}, deep learning \cite{deep} for stomach abnormality classification, etc.

In the paper by Escober et al.~\cite{escobar2021accurate}, they provided a method for classifying illnesses and abnormalities of the gastrointestinal tract in endoscopic pictures that outperformed existing approaches. The suggested technique is primarily focused on transfer learning via VGG16 convolutional neural network, which had previously been trained using the ImageNet dataset. CNNs \cite{cnns,cnn1}  have a number of distinct hidden layers, and one of their strongest skills is learning hierarchical concept representation layers that match to various degrees of abstraction. These networks perform best when the weights that fundamentally determine how the network operates are calculated using huge data. Unfortunately, because it is a costly operation, these big data sets are typically not accessible in the medical profession. Due to this, the authors proposed a transfer learning method for detecting gastrointestinal abnormalities and disorders in endoscopic images using the VGG16 \cite{vgg} CNN which had already been trained using the ImageNet dataset.

Alexey Dosovitskiy et al.~\cite{dosovitskiy2020image} looked into how Transformers might be used directly for image classification. They have developed a method for creating images as a series of patches that is then processed by a common Transformer encoder used in NLP. When combined with pre-training on substantial datasets, this method performs quite well. Vision Transformer (ViT) performs exceptionally well when the computational complexity of pre-training the model is taken into account, reaching the final state on most reduced pre-training cost. As a result, Vision Transformer is reasonably inexpensive to pre-train and meets or outperforms on numerous image classification datasets.

Scaling Vision Transformer~\cite{zhai2022scaling} claims that huge models utilize high computation resources more effectively in addition to performing better with appropriate scaling of Transformers in NLP. Understanding a model's scaling features is essential to properly developing subsequent generations since scale is a vital component in achieving outstanding outcomes. For ViT models with sufficient training data, the efficiency compute frontier typically resembles a power law. Importantly, in order to remain on this, one must concurrently scale computation and model capacity. If it fails to do so then additional compute becomes available which is not the best course of action.

Vision Transformers with Patch Diversification~\cite{gong2021vision} utilized special loss algorithms in vision transformer training to successfully promote diversity among patch representations for enhanced discriminative feature extraction. Because it enables for training to be stabilized, we can now develop vision transformers that are wider and deeper. We could improve vision transformer performance by modifying the transformer architecture to include convolution layers. Data loss and performance loss occur as a result of the self-centered blocks' preference to map different patches into equivalent latent models for visual transformers. Furthermore, without changing the transformer model structure, it is possible to train larger, deeper models and enhance performance on picture classification tasks by diversifying patch representations.

\section{Methodology}

\begin{figure}[htp]
    \centering
    \includegraphics[width=8cm]{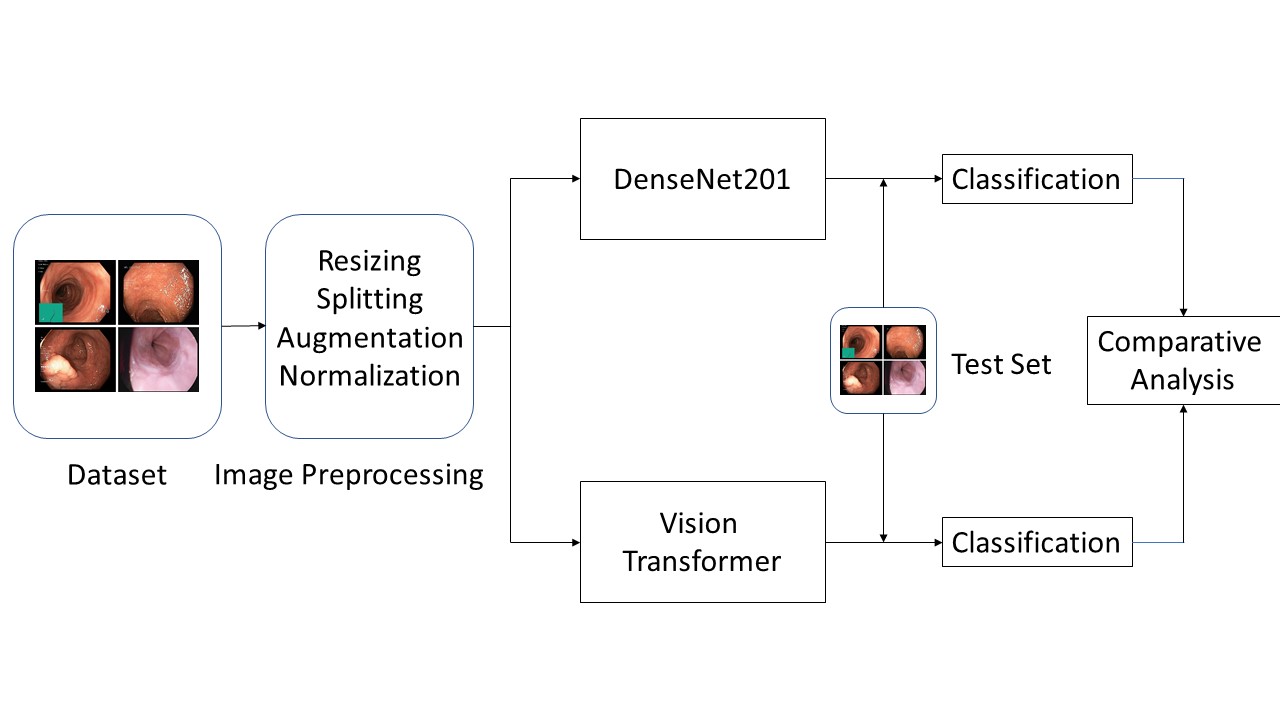}
    \caption{Proposed gastrointestinal disease detection framework using ViT and DenseNet}
    \label{fig:method}
\end{figure}

In this paper, we have proposed a novel framework to detect gastrointestinal diseases from wireless capsule endoscopy (WCE) curated images with vision transformer (ViT) based model, and pretrained DenseNet201 \cite{den}. The proposed framework is depicted in Fig. \ref{fig:method}.

\subsection{Dataset Description}

\begin{figure}[htp]
    \centering
    \includegraphics[width=8cm]{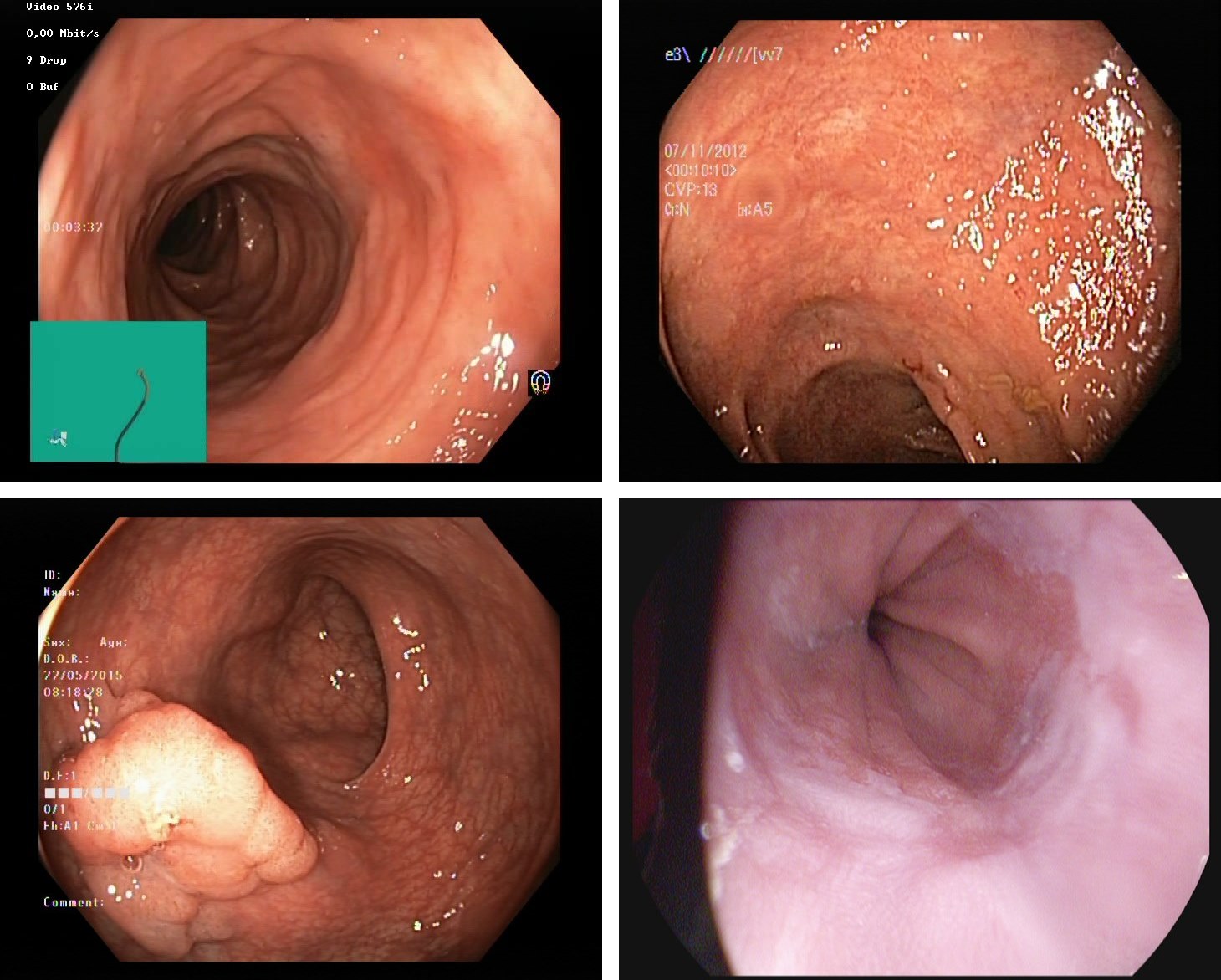}
    \caption{Sample images from dataset. Top left is a normal colon image, top right is a ulcerative colitis diseased colon image, bottom left is a polyps, and bottom right is esophagitis diseased colon WCE image}
    \label{fig:sample}
\end{figure}

\begin{figure}[htp]
    \centering
    \includegraphics[width=8cm]{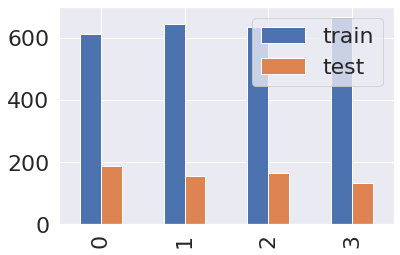}
    \caption{Distribution of training and test data in four classes. Validation data are included in training data.}
    \label{fig:dist}
\end{figure}

We have collected our dataset from Kaggle~\cite{montalbo_2022}. The dataset contained WCE images from inside the gastrointestinal (GI) tract. This dataset originally contained photos of 720 x 576 pixels of four classes: normal, ulcerative colitis, polyps, and esophagitis. We have used our machine learning models to classify this dataset into above mentioned four classes. Sample images from dataset is presented in Fig. \ref{fig:sample}. Training and test data distribution is presented in Fig. \ref{fig:dist}.

\subsection{Dataset Preprocessing}
We have divided the dataset into three sets: training set, validation set, and test set, to train and test our models on various quantitative performance evaluation metrics. We have utilized scikit library to split the dataset into training, test, and validation set. The dataset was splitted into training and test set in the ration of 8:2. Further, the training set was again split into training set and validation set in 9:1 ratio. We have resized our images to 72x72x3 for Vision Transformer based model, and 224x224x3 for DenseNet201. Further, we have labeled the classes with numerical values for the models to classify. Normal, ulcerative colitis, polyps, and esophagitis were labeled with `0', `1', `2', and `3' respectively. We have normalized the pixel values of images by dividing them by 255. To overcome data limitation, we have augmented our dataset images by Keras' ImageDataGenerator function.

\subsection{Model Architecture}

\subsubsection{DenseNet201}
\begin{figure}[htp]
    \centering
    \includegraphics[width=8cm]{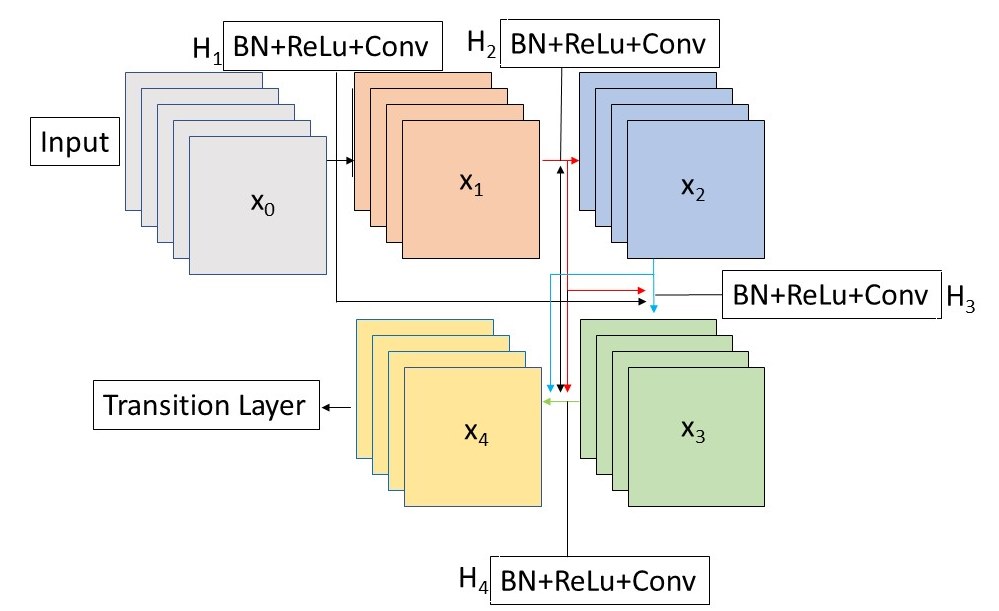}
    \caption{Deep layers of DenseNet201}
    \label{fig:deep_layers}
\end{figure}

Dense Convolutional Network or DenseNet201~\cite{huang2017densely} was developed by Huang et al. It is a pretrained feed-forward convolutional neural network model where each layer is connected to every other layer. It has L(L+1)/2 connections, where L is the number of layers. In traditional convolutional neural network, L number of layers have L connections. Authors trained it on four benchmark tasks namely, CIFAR-10,
CIFAR-100, SVHN, and ImageNet. In DenseNe architecture, $l^{th}$ layer gets feature maps from all the layers before it. Feature map of $l^{th}$ layer is defined by, 

\[x_{l} = H_{l}([x_{0},x_{1}...,x_{l-1}])\]

where, $x_{l}$ is the feature map of $l^{th}$ layer, $[x_{0},x_{1}...,x_{l-1}]$ is feature maps produced in $0,..,l-1$ layers, and $H_{l}$ is non-linear transformation function. A five layer dense block is depicted in Fig. \ref{fig:deep_layers}.

\begin{figure}[htp]
    \centering
    \includegraphics[width=8cm]{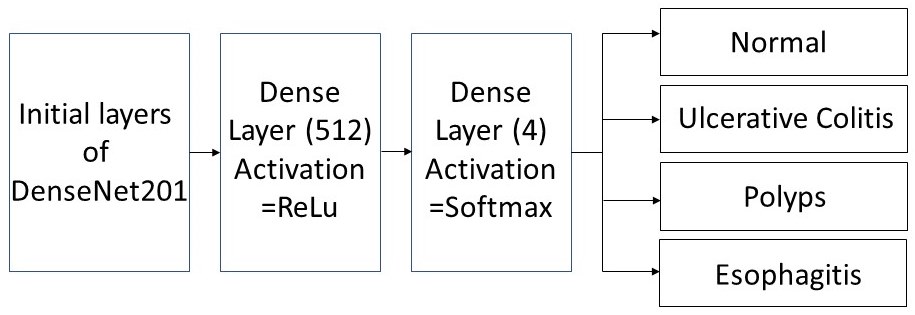}
    \caption{Overview of custom DenseNet201 architecture}
    \label{fig:dn_archi}
\end{figure}

As our classification task consists of four classes, we removed the top layer of DenseNet201 and added one dense layer consisting of 512 neurons with relu activation function and one output layer of four neurons with softmax activation function. We have used softmax activation function in the output layer as our task is a multiclass classification task. A brief overview of the model is presented in Fig. \ref{fig:dn_archi}. Our custom DenseNet201 model contains a total of 66,493,508 parameters, of which 66,264,452 are trainable and 229,056 are non trainable. ImageNet Large-Scale Visual Recognition (ILSVRC) dataset~\cite{russakovsky2015imagenet} pretrained weights are used in this model. To train our model, we have used Adam optimizer with a learning rate of 0.00001, categorical cross entropy as our loss function. We have trained our model for 100 epochs and used early stopping function to avoid overfitting of the model.

\subsubsection{Vision Transformer}
\begin{figure}[htp]
    \centering
    \includegraphics[width=8cm]{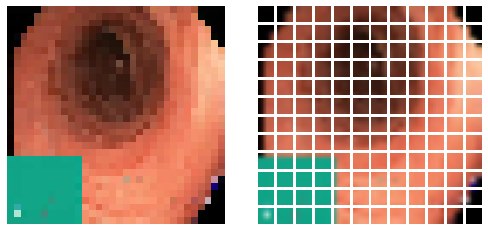}
    \caption{Vision transformer patches of a sample image}
    \label{fig:patch}
\end{figure}

Although transformers were predominantly used in natural language processing, Developed by Alexey Dosovitskiy et al.~\cite{dosovitskiy2020image} showed pure transformer based approach to classify images can be a efficient alternative to traditional CNN models. They trained vision transformers (ViT) on image recognition benchmarks such as, ImageNet, CIFAR-100, VTAB, etc. and showed their high accuracy with significantly lower computational resources to train the models. This pretrained supervised machine learning algorithm splits images into patches and applies keys/tokens to those patches, similar to the working principal of transformers in natural language processing. A depiction of image from our training set divided into patches for training vision transformer is showed in Fig. \ref{fig:patch}.

\begin{figure}[htp]
    \centering
    \includegraphics[width=8cm]{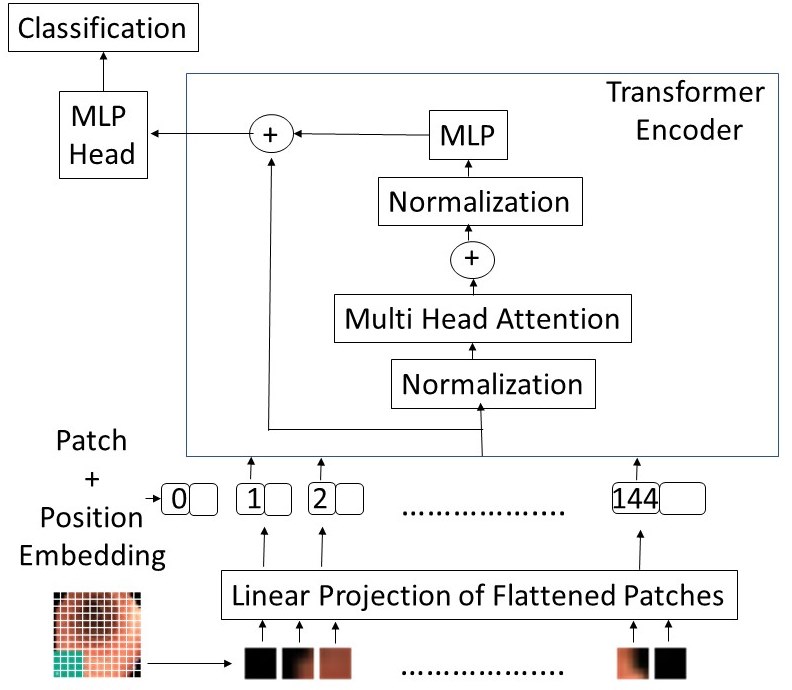}
    \caption{Overview of vision transformer architecture}
    \label{fig:vit_archi}
\end{figure}

In vision transformer based model, a 2D image $x \epsilon \mathbb{R}^{H\times W\times C}$ is reshaped into flattened sequence of 2D patches $x_p \epsilon \mathbb{R}^{N*(P^{2}.c)}$. Here, where (H, W) = original image resolution, C = number of channels, (P, P) = image patch resolution. Number of patches is given by $N = HW/P^2$. N is also the length of transformer~\cite{dosovitskiy2020image}. Positional embeddings are assigned to the patches and each sequence of patches are attached with learnable embeddings. A brief overview of vision transformer based classification model used in our work is depicted in Fig. \ref{fig:vit_archi}. Multi headed self attention block and multi layer perceptrons blocks are applied alternatively in transformer encoder. Layernorm and residual connections are applied before and after every block~\cite{dosovitskiy2020image}. In our model, we have used patch size of 6 X 6, 144 patches per image, and 108 elements per image. Parameters used in our model is shown in table \ref{tab:vit_param}.

\begin{table}[]
\centering
\caption{Parameters of vision transformer based classifier}
\label{tab:vit_param}
\begin{tabular}{|l|l|}
\hline
Parameters         & Values      \\ \hline
Learning rate      & 0.0001      \\ \hline
Weight decay       & 0.0001      \\ \hline
Batch size         & 256         \\ \hline
Number of epochs   & 100         \\ \hline
Image size         & 72x72       \\ \hline
Patch size         & 6x6         \\ \hline
Number of patches  & 144         \\ \hline
Transformer layers & 8           \\ \hline
Projection\_dim    & 64          \\ \hline
Transformer size   & 64*2, 64    \\ \hline
MLP head           & (2042,1048) \\ \hline
\end{tabular}
\end{table}

\section{Result Analysis}

We have compared our models on different quantitative evaluation metrics. To quantify the performances of our models, we have tested them on our test set which was completely unseen to them. The metrics that were used to evaluate their performance are: accuracy (\ref{eq2}), precision (\ref{eq3}), recall (\ref{eq4}), and f1 score (\ref{eq5}). All these parameters were measured on test set. A comparative analysis on the models' performances on these parameters are presented in table \ref{tab:comparison}.

\begin{equation} \label{eq2}
\begin{split}
Accuracy & = \frac{TN + TP}{TP + FP + TN+ FN} \\
\end{split}
\end{equation}
Here, TN = True negative, TP = True positive, FN = False negative, FP = False positive.

\begin{equation} \label{eq3}
\begin{split}
Precision & = \frac{TP}{TP + FP} \\
\end{split}
\end{equation}
Here, TP = True positive, FP = False positive.

\begin{equation} \label{eq4}
\begin{split}
Recall & = \frac{TP}{TP + FN} \\
\end{split}
\end{equation}
Here, TP = True positive, FN = False negative.

\begin{equation} \label{eq5}
\begin{split}
F1 Score & = \frac{2*Precision*Recall}{Precision + Recall} \\
\end{split}
\end{equation}

\begin{table}[]
\centering
\caption{A comparison table between Vision Transformer (ViT) and DenseNet201 on performance evaluation metrics}
\label{tab:comparison}
\begin{tabular}{|l|l|l|l|}
\hline
Parameters & Class                                                                                       & DenseNet201                                                               & ViT                                                       \\ \hline
Accuracy   &                                                                                             & 95.63\%                                                           & 71.88\%                                                           \\ \hline
Precision  & \begin{tabular}[c]{@{}l@{}}Normal\\ Ulcerative Colitis\\ Polyps\\ Esophagitis\end{tabular}  & \begin{tabular}[c]{@{}l@{}}1.00\\ 1.00\\ 0.61\\ 0.97\end{tabular} & \begin{tabular}[c]{@{}l@{}}0.99\\ 0.98\\ 0.71\\ 0.99\end{tabular} \\ \hline
Recall     & \begin{tabular}[c]{@{}l@{}}Normal \\ Ulcerative Colitis\\ Polyps\\ Esophagitis\end{tabular} & \begin{tabular}[c]{@{}l@{}}1.00\\ 0.32\\ 0.99\\ 1.00\end{tabular} & \begin{tabular}[c]{@{}l@{}}1.00\\ 0.60\\ 0.97\\ 1.00\end{tabular} \\ \hline
F1 Score   & \begin{tabular}[c]{@{}l@{}}Normal\\ Ulcerative Colitis\\ Polyps\\ Esophagitis\end{tabular}  & \begin{tabular}[c]{@{}l@{}}1.00\\ 0.48\\ 0.76\\ 0.99\end{tabular} & \begin{tabular}[c]{@{}l@{}}1.00\\ 0.74\\ 0.82\\ 0.99\end{tabular} \\ \hline
\end{tabular}
\end{table}

From Table \ref{tab:comparison} we can see that ViT based model outperformed DenseNet201 in test accuracy, where ViT scored 23.75\% higher than DenseNet201.

In terms of precision, DenseNet201 scored 0.01 and 0.02 more than Vit in classifying normal and ulceratice colitis images. But in classifying polyps, ViT scored significantly higher, about 0.2 and .02 higher than DenseNet201. 

In terms of recall, DenseNet201 and ViT scored same in classifying normal images, whereas, ViT scored 0.28 higher than DenseNet201 in classifying ulcerative colitis. Both the model scored same in esophagitis class. But DenseNet201 scored 0.02 higher than VitT in classifying polyps images.

In f1 scoring, both DenseNet201 and ViT scored 1.0 and 0.99 in classifying normal and ulcerative colitis images. On the other hand, ViT scored 0.26, 0.06 higher than DenseNet201 in classifying polyps and esophagitis images.

\begin{figure}[htp]
    \centering
    \includegraphics[width=8cm]{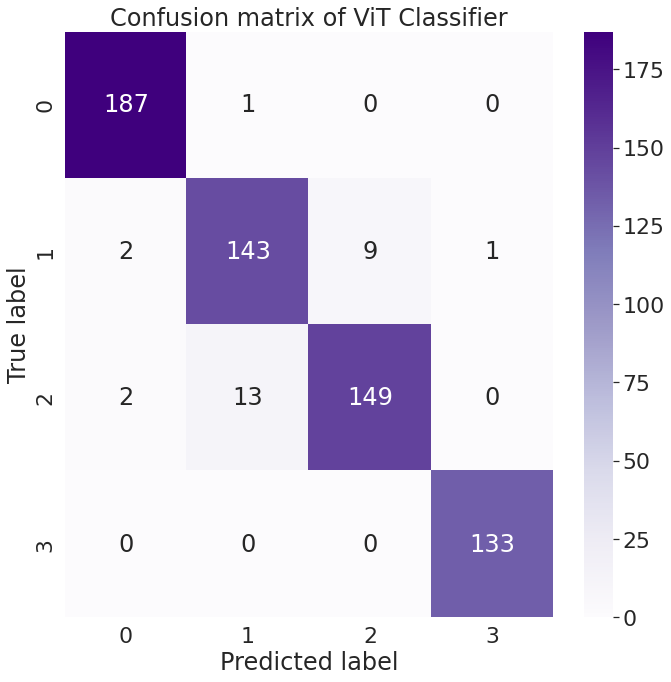}
    \caption{Confusion matrix of Vision Transformer (ViT) based model}
    \label{fig:vit_cm}
\end{figure}

\begin{figure}[htp]
    \centering
    \includegraphics[width=8cm]{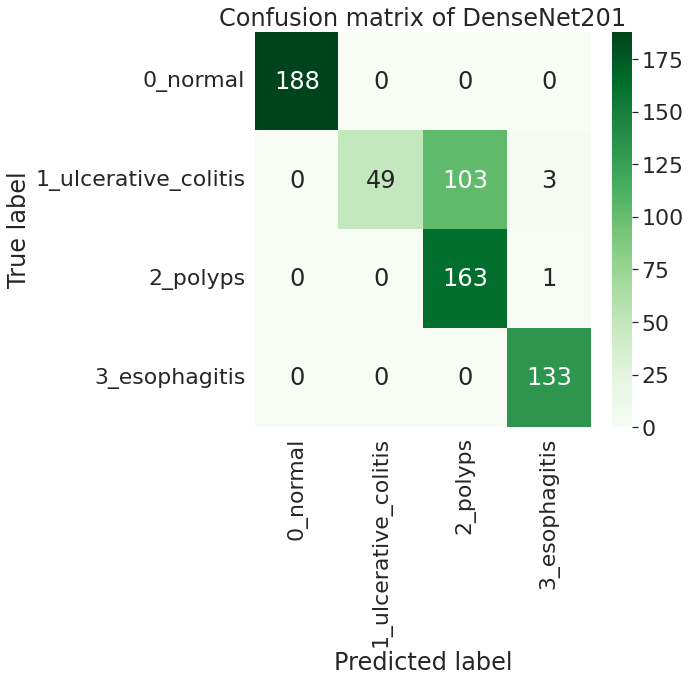}
    \caption{Confusion matrix of DenseNet201}
    \label{fig:dn_cm}
\end{figure}

We have also plotted confusion matrix of ViT and DenseNet201's performance conducted on our test set which are depicted in Fig. \ref{fig:vit_cm} and Fig. \ref{fig:dn_cm} respectively. 

From Fig. \ref{fig:vit_cm} we can see that ViT classifier detected 187 normal images correctly but misclassified one normal image. Whereas, DenseNet201 successfully classified 188 normal images. In case of Ulcerative colitis images, ViT classifier correctly classified 143 images out of 155. DenseNet201 correctly classified 103 ulcerative colitis images out of 155. ViT Classifier also could correctly classify 149 polyps images out of 164 whereas, in the case of DenseNet201 the number of correctly classified polyps images were 163 out of 164. Both ViT and DenseNet201 could correctly classify all the esophagus images in their test set.

\begin{figure}[htp]
    \centering
    \includegraphics[width=8cm]{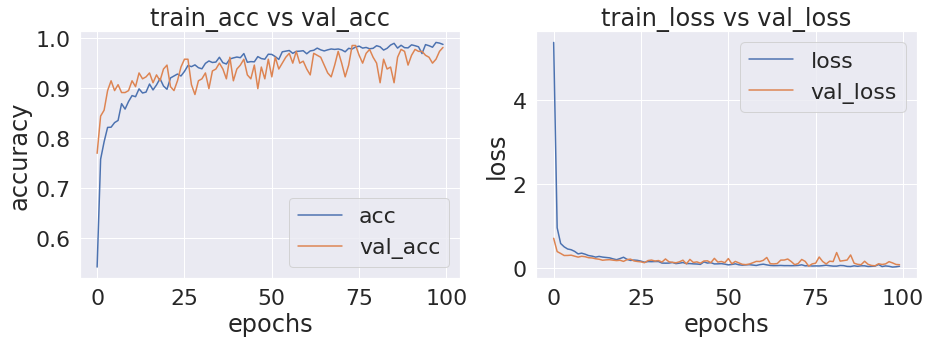}
    \caption{Accuracy and loss of Vision Transformer based model on test dataset}
    \label{fig:acc}
\end{figure}

\begin{figure}[htp]
    \centering
    \includegraphics[width=8cm]{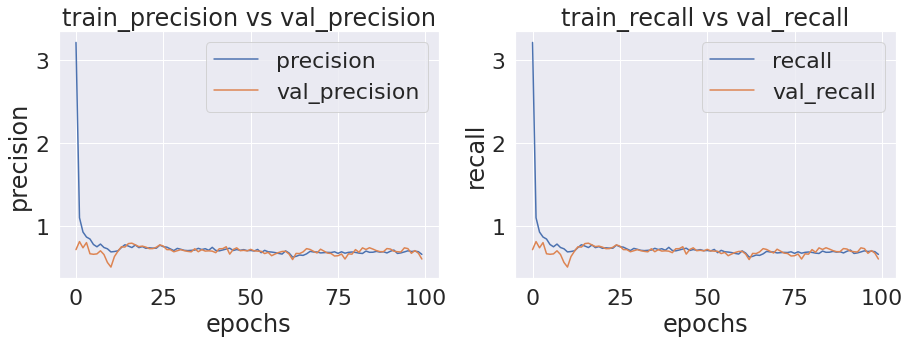}
    \caption{Precision and recall of Vision Transformer based model on test dataset}
    \label{fig:prec}
\end{figure}

Accuracy, loss, precision, and recall graph of ViT on test data is depicted in Fig. \ref{fig:acc}, and Fig. \ref{fig:prec}.

\section{Conclusion}
In this paper, we have used transfer learning approach with DenseNet201 and Vision Transformer based architecture to detect three gastrointestinal diseases: ulcerative colitis, polyps, and esophagitis, along with healthy colon images. Among the two models, Vision Transformer outperformed DenseNet201 with an accuracy of 95.63\%, whereas in the case of DenseNet201 it was 71.88\%. We have faced resource utility and data limitation in conducting our work. We resolved data limitation by adopting augmentation approach. We aim to resolve our infrastructural limitation in near future. In future we plan to work on larger range of gastrointestinal diseases with improved accuracy of vision transformer based approach. We aim to further ease the diagnosis with elastography technique using ultrasound~\cite{kabir2016improved},~\cite{7835370}.

\bibliographystyle{IEEEtran}
\bibliography{1cite}

\begin{thebibliography}{10}
\providecommand{\url}[1]{#1}
\csname url@samestyle\endcsname
\providecommand{\newblock}{\relax}
\providecommand{\bibinfo}[2]{#2}
\providecommand{\BIBentrySTDinterwordspacing}{\spaceskip=0pt\relax}
\providecommand{\BIBentryALTinterwordstretchfactor}{4}
\providecommand{\BIBentryALTinterwordspacing}{\spaceskip=\fontdimen2\font plus
\BIBentryALTinterwordstretchfactor\fontdimen3\font minus
  \fontdimen4\font\relax}
\providecommand{\BIBforeignlanguage}[2]{{%
\expandafter\ifx\csname l@#1\endcsname\relax
\typeout{** WARNING: IEEEtran.bst: No hyphenation pattern has been}%
\typeout{** loaded for the language `#1'. Using the pattern for}%
\typeout{** the default language instead.}%
\else
\language=\csname l@#1\endcsname
\fi
#2}}
\providecommand{\BIBdecl}{\relax}
\BIBdecl

\bibitem{Haile_Salau_Enyew_Belay_2022}
\BIBentryALTinterwordspacing
M.~B. Haile, A.~O. Salau, B.~Enyew, and A.~J. Belay,
  ``\BIBforeignlanguage{en}{Detection and classification of gastrointestinal
  disease using convolutional neural network and svm},'' Jun 2022. [Online].
  Available: \url{http://dx.doi.org/10.1080/23311916.2022.2084878}
\BIBentrySTDinterwordspacing

\bibitem{Sitarz_Skierucha_Mielko_Offerhaus_Maciejewski_Polkowski_2018}
\BIBentryALTinterwordspacing
R.~Sitarz, M.~Skierucha, J.~Mielko, J.~Offerhaus, R.~Maciejewski, and
  W.~Polkowski, ``\BIBforeignlanguage{en}{Gastric cancer: epidemiology,
  prevention, classification, and treatment},'' p. 239â248, Feb 2018.
  [Online]. Available: \url{http://dx.doi.org/10.2147/CMAR.S149619}
\BIBentrySTDinterwordspacing

\bibitem{Tan_2013}
\BIBentryALTinterwordspacing
C.~L. Tan, ``\BIBforeignlanguage{en}{Muco-submucosal elongated polyps of the
  gastrointestinal tract: A case series and a review of the literature},'' p.
  1845, 2013. [Online]. Available:
  \url{http://dx.doi.org/10.3748/wjg.v19.i11.1845}
\BIBentrySTDinterwordspacing

\bibitem{Moayyedi_Santana_Khan_Preston_Donnellan_2007}
\BIBentryALTinterwordspacing
P.~Moayyedi, J.~Santana, M.~Khan, C.~Preston, and C.~Donnellan, ``Medical
  treatments in the short term management of reflux oesophagitis,'' Apr 2007.
  [Online]. Available: \url{http://dx.doi.org/10.1002/14651858.CD003244.pub2}
\BIBentrySTDinterwordspacing

\bibitem{JHA2021102007}
\BIBentryALTinterwordspacing
D.~Jha, S.~Ali, S.~Hicks, V.~Thambawita, H.~Borgli, P.~H. Smedsrud, T.~{de
  Lange}, K.~Pogorelov, X.~Wang, P.~Harzig, M.-T. Tran, W.~Meng, T.-H. Hoang,
  D.~Dias, T.~H. Ko, T.~Agrawal, O.~Ostroukhova, Z.~Khan, M.~{Atif Tahir},
  Y.~Liu, Y.~Chang, M.~KirkerÃžd, D.~Johansen, M.~Lux, H.~D. Johansen, M.~A.
  Riegler, and P.~Halvorsen, ``A comprehensive analysis of classification
  methods in gastrointestinal endoscopy imaging,'' \emph{Medical Image
  Analysis}, vol.~70, p. 102007, 2021. [Online]. Available:
  \url{https://www.sciencedirect.com/science/article/pii/S1361841521000530}
\BIBentrySTDinterwordspacing

\bibitem{Ozturk_Ozkaya_2021}
\BIBentryALTinterwordspacing
Å.~ÃztÃŒrk and U.~Ãzkaya, ``\BIBforeignlanguage{en}{Residual lstm layered cnn
  for classification of gastrointestinal tract diseases},'' p. 103638, Jan
  2021. [Online]. Available: \url{http://dx.doi.org/10.1016/j.jbi.2020.103638}
\BIBentrySTDinterwordspacing

\bibitem{Majid_Khan_Yasmin_Rehman_Yousafzai_Tariq_2020}
\BIBentryALTinterwordspacing
A.~Majid, M.~A. Khan, M.~Yasmin, A.~Rehman, A.~Yousafzai, and U.~Tariq,
  ``\BIBforeignlanguage{en}{Classification of stomach infections: A paradigm of
  convolutional neural network along with classical features fusion and
  selection},'' p. 562â576, Jan 2020. [Online]. Available:
  \url{http://dx.doi.org/10.1002/jemt.23447}
\BIBentrySTDinterwordspacing

\bibitem{deep}
K.~M. Hasib, M.~A. Habib, N.~A. Towhid, and M.~I.~H. Showrov, ``A novel deep
  learning based sentiment analysis of twitter data for us airline service,''
  in \emph{2021 International Conference on Information and Communication
  Technology for Sustainable Development (ICICT4SD)}, 2021, pp. 450--455.

\bibitem{escobar2021accurate}
J.~Escobar, K.~Sanchez, C.~Hinojosa, H.~Arguello, and S.~Castillo, ``Accurate
  deep learning-based gastrointestinal disease classification via transfer
  learning strategy,'' in \emph{2021 XXIII Symposium on Image, Signal
  Processing and Artificial Vision (STSIVA)}.\hskip 1em plus 0.5em minus
  0.4em\relax IEEE, 2021, pp. 1--5.

\bibitem{cnns}
K.~M. Hasib, N.~A. Towhid, and M.~G.~R. Alam, ``Online review based sentiment
  classification on bangladesh airline service using supervised learning,'' in
  \emph{2021 5th International Conference on Electrical Engineering and
  Information Communication Technology (ICEEICT)}, 2021, pp. 1--6.

\bibitem{cnn1}
S.~Ahmed, M.~H.~K. Mehedi, M.~Rahman, and J.~Bin~Sayed, ``Bangla music lyrics
  classification,'' 09 2022, pp. 142--147.

\bibitem{vgg}
\BIBentryALTinterwordspacing
S.~Montaha, S.~Azam, A.~K. M. R.~H. Rafid, M.~Z. Hasan, A.~Karim, K.~M. Hasib,
  S.~K. Patel, M.~Jonkman, and Z.~I. Mannan, ``Mnet-10: A robust shallow
  convolutional neural network model performing ablation study on medical
  images assessing the effectiveness of applying optimal data augmentation
  technique,'' \emph{Frontiers in Medicine}, vol.~9, 2022. [Online]. Available:
  \url{https://www.frontiersin.org/articles/10.3389/fmed.2022.924979}
\BIBentrySTDinterwordspacing

\bibitem{dosovitskiy2020image}
A.~Dosovitskiy, L.~Beyer, A.~Kolesnikov, D.~Weissenborn, X.~Zhai,
  T.~Unterthiner, M.~Dehghani, M.~Minderer, G.~Heigold, S.~Gelly \emph{et~al.},
  ``An image is worth 16x16 words: Transformers for image recognition at
  scale,'' \emph{arXiv preprint arXiv:2010.11929}, 2020.

\bibitem{zhai2022scaling}
X.~Zhai, A.~Kolesnikov, N.~Houlsby, and L.~Beyer, ``Scaling vision
  transformers,'' in \emph{Proceedings of the IEEE/CVF Conference on Computer
  Vision and Pattern Recognition}, 2022, pp. 12\,104--12\,113.

\bibitem{gong2021vision}
C.~Gong, D.~Wang, M.~Li, V.~Chandra, and Q.~Liu, ``Vision transformers with
  patch diversification,'' \emph{arXiv preprint arXiv:2104.12753}, 2021.

\bibitem{den}
K.~M. Hasib, S.~Sakib, J.~A. Mahmud, K.~Mithu, M.~S. Rahman, and M.~S. Alam,
  ``Covid-19 prediction based on infected cases and deaths of bangladesh using
  deep transfer learning,'' in \emph{2022 IEEE World AI IoT Congress (AIIoT)},
  2022, pp. 296--302.

\bibitem{montalbo_2022}
\BIBentryALTinterwordspacing
F.~J. Montalbo, ``Wce curated colon disease dataset deep learning,'' Apr 2022.
  [Online]. Available:
  \url{https://www.kaggle.com/datasets/francismon/curated-colon-dataset-for-deep-learning}
\BIBentrySTDinterwordspacing

\bibitem{huang2017densely}
G.~Huang, Z.~Liu, L.~Van Der~Maaten, and K.~Q. Weinberger, ``Densely connected
  convolutional networks,'' in \emph{Proceedings of the IEEE conference on
  computer vision and pattern recognition}, 2017, pp. 4700--4708.

\bibitem{russakovsky2015imagenet}
O.~Russakovsky, J.~Deng, H.~Su, J.~Krause, S.~Satheesh, S.~Ma, Z.~Huang,
  A.~Karpathy, A.~Khosla, M.~Bernstein \emph{et~al.}, ``Imagenet large scale
  visual recognition challenge,'' \emph{International journal of computer
  vision}, vol. 115, no.~3, pp. 211--252, 2015.

\bibitem{kabir2016improved}
I.~E. Kabir, R.~Abid, A.~S. Ashik, K.~K. Islam, and S.~K. Alam, ``Improved
  strain estimation using a novel 1.5 d approach: Preliminary results,'' in
  \emph{2016 International Conference on Medical Engineering, Health
  Informatics and Technology (MediTec)}.\hskip 1em plus 0.5em minus 0.4em\relax
  IEEE, 2016, pp. 1--5.

\bibitem{7835370}
R.~A. Mukaddim, J.~Shan, I.~E. Kabir, A.~S. Ashik, R.~Abid, Z.~Yan, D.~N.
  Metaxas, B.~S. Garra, K.~K. Islam, and S.~K. Alam, ``A novel and robust
  automatic seed point selection method for breast ultrasound images,'' in
  \emph{2016 International Conference on Medical Engineering, Health
  Informatics and Technology (MediTec)}, 2016, pp. 1--5.

\end{thebibliography}
\end{document}